\documentclass[11pt,a4paper]{article}
\usepackage[hyperref]{emnlp2020}
\usepackage{times}
\usepackage{latexsym}

\usepackage{url}
\usepackage{graphicx}
\usepackage{verbatim}
\usepackage{multirow}
\usepackage{courier}

\usepackage{microtype}

\aclfinalcopy 

\newcommand{\criteria}{31}
\newcommand{\tools}{93}

\title{Annotationsaurus: A Searchable Directory of Annotation Tools}

\author{
  Mariana Neves \\
  German Federal Institute for \\
  Risk Assessment (BfR), \\
  German Center for the Protection \\
  of Laboratory Animals (Bf3R), \\
  Berlin, Germany \\
  {\tt \small{mariana-lara.neves@bfr.bund.de}} \\
  \And
  Jurica \v{S}eva \\
  Ada Health GmbH, \\
  Karl-Liebknecht-Str. 1, \\
  10178 Berlin, Germany \\
  {\tt \small{jurica.seva@ada.com}} \\
 }

\date{}

\begin{document}
\maketitle

\begin{abstract}

Manual annotation of textual documents is a necessary task when constructing benchmark corpora for training and evaluating machine learning algorithms.
We created a comprehensive directory of annotation tools that currently includes \tools{} tools.
We analyzed the tools over a set of \criteria{} features and implemented simple scripts and a Web application that filters the tools based on chosen criteria.
We present two use cases using the directory and propose ideas for its maintenance.
The directory, source codes for scripts, and link to the Web application are available at: \url{https://github.com/mariananeves/annotation-tools}

\end{abstract}

\section{Introduction}

Annotation tools are important resources for building customized corpora for the most various purposes, ranging from document classification \cite{btv585} to discourse relations \cite{P18-2071}.
When choosing an annotation tool for a particular project, researchers should consider the task at hand, the particularities of their system environment and the expertise of the annotators.
Frequently, it is necessary to experiment with tools, even with the burden of installing them, before finding the one that best fits their needs.

This is, however, a demanding task, given the high number of available tools.
Usually, while choosing an annotation tool, users ask for advice in forums and mailing lists, look at previous surveys of annotations tools, e.g. \cite{10.1093/bib/bbz130}, \cite{10.1093/bib/bbs084} and \cite{Fort:2016:CAR:3056070}, and check the tools that have been used in recently published corpora.
As far as we know, there is no comprehensive and searchable directory of annotation tools.

We recently defined a set of \criteria{} criteria and carried out hands-on experiments with 15 selected annotation tools from a total of \tools{} tools \cite{10.1093/bib/bbz130}.
As an extension to this survey, we now created a directory of annotation tools, evaluated the remaining (non-selected) tools based on many of these criteria and developed scripts and a Web application to allow searching for tools.
In the next section we describe our directory, discuss how to search for annotation tools, and present two use cases.

\section{Directory of annotation tools}

We created a directory of annotation tools and evaluated the tools over some desirable criteria.
This work considers the tools that we previously collected during the preparation of a survey \cite{10.1093/bib/bbz130}, 
extended with the ones that we have recently found.
Currently, the directory contains a total of \tools{} tools, as listed in 
our GitHub repository\footnote{\url{https://github.com/mariananeves/annotation-tools}}, along with links to the corresponding publication (if any) and tool's Web site (if found).



\subsection{List of attributes}

We considered both the so-called requirements and criteria (hereafter called only as criteria) that we defined in our survey \cite{10.1093/bib/bbz130}.
There, we defined five requirements for selecting tools for hands-on experiments and 26 criteria which we used for the evaluation.
The criteria are related to the tool's publications, technical attributes, data format, and functional properties (cf. Table~\ref{tab:criteria}). 
The chosen  \criteria{} criteria, including the possible values that each criterion can take, as defined in our survey, is available in our GitHub repository\footnote{\url{https://github.com/mariananeves/annotation-tools/blob/master/schema}}. 

\begin{table*}
\centering
\begin{tabular}{c|lll}
\multicolumn{2}{c}{Criteria} & Values & Filter \\
\hline
\parbox[t]{3mm}{\multirow{5}{*}{\rotatebox[origin=lc]{90}{\#requirements}}}
& Availability of the tool & yes,no & available \\
& Type of installation & web-based,stand-alone,plug-in & type \\
& Installation successful & yes,no & installable \\
& Annotation successful & yes,no & workable \\
& Ability to configure a schema & yes,no & schematic \\
\hline
\parbox[t]{2mm}{\multirow{3}{*}{\rotatebox[origin=c]{90}{-}}}
& Year of the last publication & - & - \\
& Number of citations for publication & - & - \\
& Number of citations for corpora & - & - \\
\hline
\parbox[t]{2mm}{\multirow{7}{*}{\rotatebox[origin=c]{90}{\#technical}}}
& Year of the last version & - & - \\
& Availability of source code & yes,no & source\_code \\
& Availability of online tool & yes,no & online\_available \\
& Easiness of installation & easy,medium,hard & installation \\
& Quality of the documentation & good,poor,none & documentation \\
& License of the tool & full,partial,none & license \\
& Free availability & yes,partial,no & free \\
\hline
\multirow{3}{*}{\rotatebox[origin=c]{90}{\parbox[t]{2mm}{\#data format}}}
& Format of the schema & XML,JSON,GUI,other & format\_schema \\
& Format i/o of documents & XML,JSON,TXT,other & format\_documents \\
& Format i/o of annotations & XML,JSON,TXT,other & format\_annotations \\
\hline
\parbox[t]{2mm}{\multirow{13}{*}{\rotatebox[origin=c]{90}{\#functional}}}
& Annotation with multiple labels & yes,no & multilabel \\
& Annotation on document level & yes,partial,no & document\_level \\
& Annotation of relations & yes,partial,no & relationships \\
& Annotation based on ontologies & yes,no & ontologies \\
& Support for pre-annotations & yes,partial,no & preannotations \\
& Integration with PubMed/PMC & yes,partial,no & medline\_pmc \\
& Support for full texts & yes,partial,no & full\_texts \\
& Support for partial saving & yes,partial,no & partial\_save \\
& Support for highlighting & yes,no & highlight \\
& Support for users and teams & yes,partial,no & users\_teams \\
& Support for IAA & yes,partial,no & iaa \\
& Annotation of private data & yes,no & data\_privacy \\
& Support for various languages & yes,partial,no & multilingual \\
\end{tabular}
\caption{List of criteria, their possible values, and corresponding filters (if available) in the Web application.}
\label{tab:criteria}
\end{table*}

\subsection{Evaluation of tools}

Besides the evaluation for the 15 tools previously considered in 
\cite{10.1093/bib/bbz130}, we also carried out an evaluation for the remaining 78 tools discarded in said survey.
However, given the large number of tools, our evaluation was restricted to the tools' publication and documentation, i.e., we could not perform hands-on experiment with all of mentioned tools.

We present each tool's evaluation (cf. ``tools'' folder) 
in a semi-structured text file (evaluation file) for each tool. 
This makes the evaluation file both human- and machine-readable.
When evaluating a tool, we created a copy of the ``schema'' file and checked each criterion by 
assigning the appropriate value, depending of it being multiple choice or not.
If we were not able to evaluate a certain criterion, we used the \# symbol in the corresponding line in the evaluation file.
Currently, we only assign one value to criteria that might accept multiple values, e.g. the data format criteria.

We obtained a total of 2,270 evaluations for all criteria over all tools, in contrast to 2,883 possible criteria (\tools{} tools x \criteria{} criteria). This discrepancy is contributed to our inability to evaluate all tools across all defined criteria.
This covers 78.7\% of all possible evaluations.

Regarding the evaluation of each criteria, the 
coverage ranges from 38 for the most exclusive criteria to \tools{} covered tools for most inclusive criteria.
Five criteria were evaluated for all tools, namely ``\textit{available}'', ``\textit{online\_available}'', ``\textit{last\_publication}'', ``\textit{citations\_corpora}'' and ``\textit{citations}''.
The criteria that we most missed were ``\textit{workable}'' (38 evaluations), ``\textit{multilingual}'' (38 evaluations), and ``\textit{installation}'' (41 evaluations).
Additionally, some of there criteria require hands-on experiments with the tools, which was not possible for all tools. 

We could evaluate all \criteria{} criteria for 15 tools, as already reported in \cite{10.1093/bib/bbz130}, and the average number of evaluated criteria was 24.
RAD was the tool for which we assessed the fewest number of criteria (only 8).
Further tools with less than 20 evaluated criteria were: Analec (13), AWOCATo (14), eHost (15), Hexatomic (16), SFA (17), OLLIE (17), Coco (18), Cas Editor (18), Slate (18), KAFnotator (18), and Serengueti (19).

\subsection{Directory maintenance}

We designed the directory focusing on the easiness of its  
maintenance.
For any modification, a user should raise an issue in the GitHub repository.
Here we discuss the possible changes that we can envisage for the repository.

\paragraph{Addition of new annotation tools.}
New annotation tools can be added simply by 
placing its evaluation file, with matching file name to the tools name, in the ``tools'' folder of the repository.
Requests for new additions can be performed by creating an issue in the GitHub repository. 
Removing a tool is then, in contrast, simply a matter of removing its evaluation file from the ``tools'' folder of the repository
We cannot, however, envisage a reason why a certain tool should be removed from the repository. 
As already stated above, we currently include all tools that we found.
Although some tools are no longer available, we include them for the sake of completeness. 

\paragraph{Addition of new criteria.}
A new criterion should first be inserted in the ``schema'' file and then replicated for each tool's evaluation file (``tools'' folder).
Adding the additional criterion (line) for each tool's file could be carried out either manually or automatically via a script, for instance, by setting a default value or leaving the corresponding criterion (line) commented out.
However, a missing criteria for a particular tool does not cause a problem for neither the scripts nor the Web tool.

\paragraph{Changes to a criterion for a tool.}
Due the number of tools and criteria, coupled with the manual nature of the evaluation, we keep the possibility of a faulty evaluation open.   
In such situations updating the evaluation is a matter of changing the appropriate line(s) in the relevant evaluation file. 
We are glad to receive feedback from the community regarding our evaluation.

\section{Scripts and Web application}

\begin{figure*}
\centering
\includegraphics[scale=0.45]{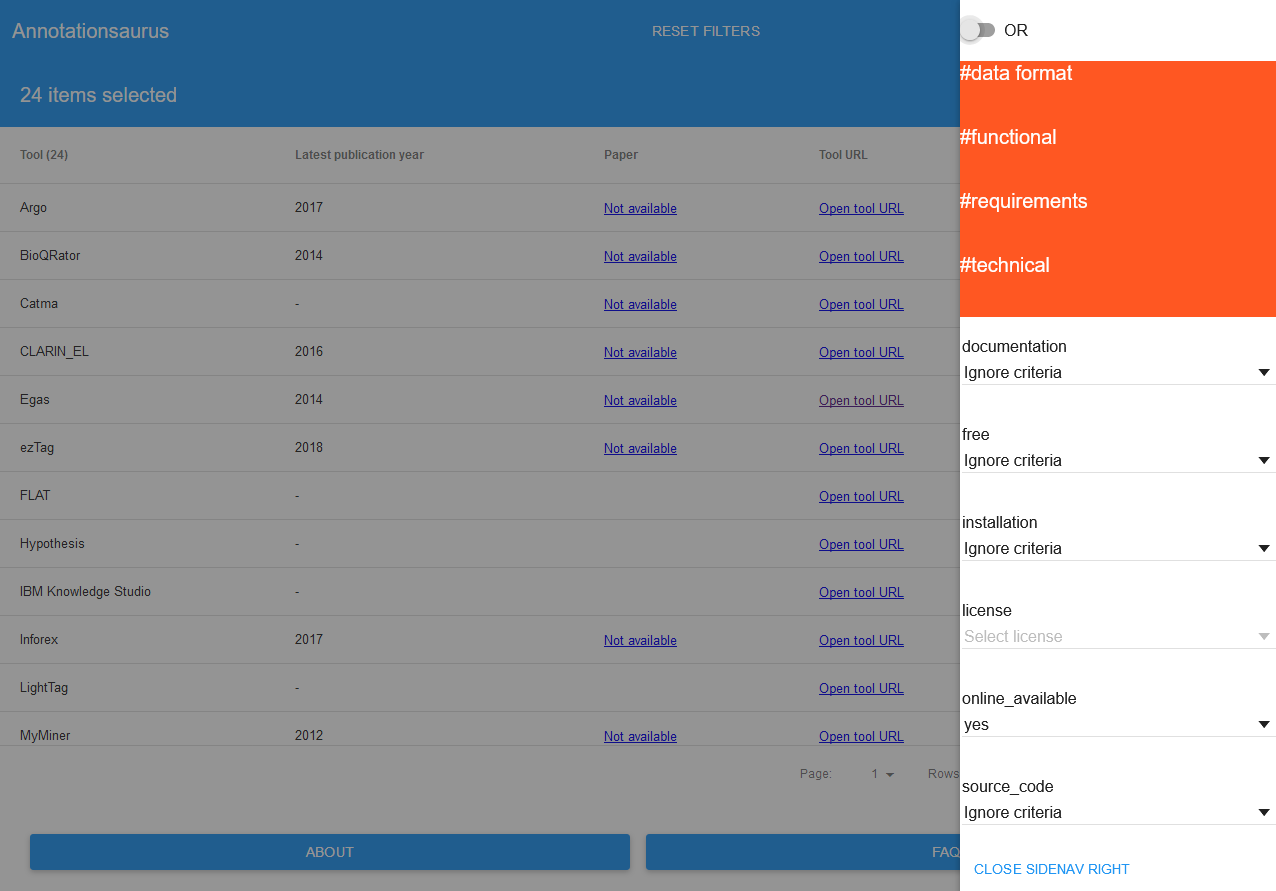}
\caption{Screen-shot of the Web application and its filters (on the right, in orange). 
}
\label{fig:webapp_filters}
\end{figure*}

We implemented a 
Web application for searching among the tools.
The system is implemented two fold. We use Python/Flask for the server side logic which consists of reading all evaluation files and serving them to the interface.
The interface was, in turn, implemented in AngularJS JavaScript framework as was designed as a responsive, single-page web application. 
It supports searching by all 
non-numerical criteria, omitting e.g. last publication year, through a dynamic side navigation bar (cf. Figure~\ref{fig:webapp_filters}). 
Initially all tools are presented in a tabular way, with configurable pagination functionality. 
By choosing at least one of the possible filters the list will be filtered to tools matching selected criteria.


The matching tools are shown in the decreasing order of the number of search criteria that were matched. 
For tools with the same number of matching criteria, alphabetical ordering is used. 
For each tool, we print the number of matched criteria and their name(s), as selected in the filtering options.
Additionally, the search can be adapted to use OR and AND truth-functional operator against the selected criteria. Initially, OR is used.
The Web application is currently running in Heroku~\footnote{\url{https://annotationsaurus.herokuapp.com/}}.

\section{Use Cases}

\begin{figure*}
\centering
\includegraphics[scale=0.58]{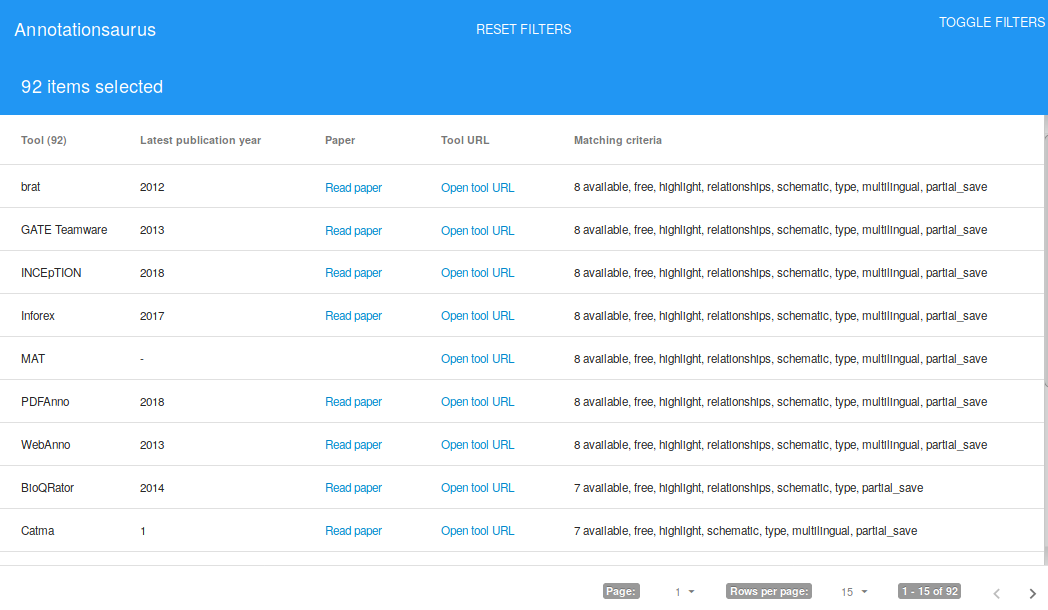}
\caption{Screen-shot of the Web application for the use case 1. We show the list of the top 10 tools returned when after selecting the filters that we defined for the use case. 
}
\label{fig:webapp}
\end{figure*}

We describe two use cases to illustrate how the repository (and search tools) could support finding the annotation tools that best fit one's needs.
For both use cases, we require annotation tools to be Web-based and freely available for research purposes (i.e. ``type=web-based'' and ``available'' in \#requirements, and ``free'' in \#technical) and that allow the definition of a schema (``schematic'' in \#requirements).
Further, we experimented with two optional criteria, namely, whether the tools are installable and workable (in \#requirements).
We did not put any limitation for the availability of source code or format for schema, documents or annotations.

\paragraph{Use Case 1: Semantic annotation of entity and relations}

Our first scenario addresses a typical annotation project of semantic annotations.
This is frequently necessary for construction of domain-specific corpora, e.g. \cite{W12-4304,W12-2426}.
This use case requires tools for manual annotation of named-entities (i.e. highlighting of text spans) and for drawing relations between these.
Therefore, we included the filters ``highlight'' and ``relationships'' (in \#functional) as mandatory.
Given the complexity and ambiguity of many domains, we expect the tool to be able to assign multiple labels to text spans (i.e. overlapping annotations), i.e., the ``multilabel'' criterion (in \#functional).
Finally, such annotation projects are demanding and might take some hours for a single (long) document, therefore, we also require the tool to be able to save documents partially (``partial-save'' criterion in \#functional).

Seven tools (brat, INCEpTION, Inforex, MAT, SANTO, TeamTat, and WebAnno) matched the eight mandatory criteria that we defined, i.e. the four general non-optional ones and the four specific ones.
Further, when checking the two general optional arguments that require the tools to be installable and workable, we had only four tools matching all filters (brat, MAT, TeamTat, and WebAnno).

\paragraph{Use Case 2: Document-level annotations for text classification}

For our second use case, we envisage the annotation of clinical data, e.g. pathology reports \cite{Jouhet2012AutomatedCO} or death certificates \cite{W16-5107}.
Thus, we require the criterion ``data-privacy'' (in \#functional) that assures that the tool can be locally installed.
Further, we address the task of annotation on the document level (criterion ``document-level'' in \#functional), such as carried out in \cite{btv585}.
As optional additional criteria, we check whether these tools provide annotations based on pre-defined ontologies, i.e. by specifying the criteria ``preannotations'' and ``ontologies'' (in \#functional).

Six tools (Bionotate, doccano, MAT, SMART, TALEN, and UniversalAnnotationTool) complied with the six mandatory filters that we defined, i.e., the four general ones and the two specific ones for the use case.
When checking the other two optional filters for this use case, no tools could comply with all eight criteria, but 12 tools matched seven of these criteria.
For instance, SANTO and WebAnno did not comply with the document-level annotation, while Bionotate and MAT do not support ontologies.

\section{Conclusions}

We described a directory of annotation tools with search functionality.
It currently includes a total of \tools{} tools that were evaluated on many of the \criteria{} criteria.
We developed a simple 
Web application to search among the tools by filtering for chosen criteria.
The tools' evaluation files and the 
Web application are available in a GitHub repository and we welcome feedback from the community regarding missing tools, interesting new criteria, and corrections of our current evaluations.

In spite of covering a total of \tools{} tools and \criteria{} criteria, our repository 
have many limitations.
Our list of annotation tools is very comprehensive, but some tools might be missing and the links to tools' publication and Web sites might become broken at any time.
Although the evaluation (plain text) files are readable by both human and scripts, this also makes the process error-prone if the criteria name is not written exactly as defined in the schema file.
Therefore, we provide a script 
for checking errors in the criteria names, as well as for printing statistics of the current state of the evaluation. 
For the sake of simplicity, we decided for a file-based approach in this first version of the directory.
Later, if necessary, these files can be converted to a more appropriate format or imported into a database.

We certainly missed some interesting criteria that could be useful for some domains or annotation projects, such as support for annotation layers, or tools suitable for annotation of messages from the social media when building specific corpora \cite{f4e6d05245c54f248fd809bcb30f28a7}.
Further, the definition of some criteria could be improved, such as specifying the corpora which where developed using a certain tool instead of simply informing how many were found.

We also envisage future work for supporting the automatic update of some criteria.
For instance, the last version (year of publication or commit) of a tool could be automatically extracted from the GitHub repository (if available), the number of citations for a publication might be automatically retrieved from Google Scholar, and the corpora which were developed using a certain tool could potentially also be automatically extracted from  publications.

\bibliographystyle{acl_natbib}
\bibliography{references}

\begin{thebibliography}{10}
\expandafter\ifx\csname natexlab\endcsname\relax\def\natexlab#1{#1}\fi

\bibitem[{Baker et~al.(2016)Baker, Silins, Guo, Ali, H\"ogberg, Stenius, and
  Korhonen}]{btv585}
Simon Baker, Ilona Silins, Yufan Guo, Imran Ali, Johan H\"ogberg, Ulla Stenius,
  and Anna Korhonen. 2016.
\newblock \href {https://doi.org/10.1093/bioinformatics/btv585} {Automatic
  semantic classification of scientific literature according to the hallmarks
  of cancer}.
\newblock \emph{Bioinformatics}, 32(3):432--440.

\bibitem[{Boyce et~al.(2012)Boyce, Gardner, and Harkema}]{W12-2426}
Richard Boyce, Gregory Gardner, and Henk Harkema. 2012.
\newblock \href {http://aclweb.org/anthology/W12-2426} {Using natural language
  processing to extract drug-drug interaction information from package
  inserts}.
\newblock In \emph{BioNLP: Proceedings of the 2012 Workshop on Biomedical
  Natural Language Processing}, pages 206--213. Association for Computational
  Linguistics.

\bibitem[{Fort(2016)}]{Fort:2016:CAR:3056070}
Kar\"en Fort. 2016.
\newblock \emph{Collaborative Annotation for Reliable Natural Language
  Processing: Technical and Sociological Aspects}, 1st edition.
\newblock Wiley-IEEE Press.

\bibitem[{Jouhet et~al.(2012)Jouhet, Defossez, Burgun, Beux, Levillain,
  Ingrand, and Claveau}]{Jouhet2012AutomatedCO}
Vianney Jouhet, Gautier Defossez, Anita. Burgun, Pierre~Le Beux, P~Levillain,
  Pierre Ingrand, and Vincent Claveau. 2012.
\newblock Automated classification of free-text pathology reports for
  registration of incident cases of cancer.
\newblock \emph{Methods of information in medicine}, 51 3:242--51.

\bibitem[{Lavergne et~al.(2016)Lavergne, Neveol, Robert, Grouin, Rey, and
  Zweigenbaum}]{W16-5107}
Thomas Lavergne, Aurelie Neveol, Aude Robert, Cyril Grouin, Gr{\'e}goire Rey,
  and Pierre Zweigenbaum. 2016.
\newblock \href {http://www.aclweb.org/anthology/W16-5107} {A dataset for
  icd-10 coding of death certificates: Creation and usage}.
\newblock In \emph{Proceedings of the Fifth Workshop on Building and Evaluating
  Resources for Biomedical Text Mining (BioTxtM2016)}, pages 60--69. The COLING
  2016 Organizing Committee.

\bibitem[{Neves and Leser(2012)}]{10.1093/bib/bbs084}
Mariana Neves and Ulf Leser. 2012.
\newblock \href {https://doi.org/10.1093/bib/bbs084} {{A survey on annotation
  tools for the biomedical literature}}.
\newblock \emph{Briefings in Bioinformatics}, 15(2):327--340.

\bibitem[{Neves and \v{S}eva(2019)}]{10.1093/bib/bbz130}
Mariana Neves and Jurica \v{S}eva. 2019.
\newblock \href {https://doi.org/10.1093/bib/bbz130} {{An extensive review of
  tools for manual annotation of documents}}.
\newblock \emph{Briefings in Bioinformatics}.
\newblock Bbz130.

\bibitem[{O'Connor et~al.(2014)O'Connor, Pimpalkhute, Nikfarjam, Ginn, Smith,
  and Gonzalez}]{f4e6d05245c54f248fd809bcb30f28a7}
Karen O'Connor, Pranoti Pimpalkhute, Azadeh Nikfarjam, Rachel Ginn, {Karen L.}
  Smith, and Graciela Gonzalez. 2014.
\newblock Pharmacovigilance on twitter? mining tweets for adverse drug
  reactions.
\newblock \emph{AMIA ... Annual Symposium proceedings / AMIA Symposium. AMIA
  Symposium}, 2014:924--933.

\bibitem[{Ohta et~al.(2012)Ohta, Pyysalo, Tsujii, and Ananiadou}]{W12-4304}
Tomoko Ohta, Sampo Pyysalo, Jun'ichi Tsujii, and Sophia Ananiadou. 2012.
\newblock \href {http://aclweb.org/anthology/W12-4304} {Open-domain anatomical
  entity mention detection}.
\newblock In \emph{Proceedings of the Workshop on Detecting Structure in
  Scholarly Discourse}, pages 27--36. Association for Computational
  Linguistics.

\bibitem[{Yang and Li(2018)}]{P18-2071}
An~Yang and Sujian Li. 2018.
\newblock \href {http://aclweb.org/anthology/P18-2071} {{SciDTB}: Discourse
  dependency treebank for scientific abstracts}.
\newblock In \emph{Proceedings of the 56th Annual Meeting of the Association
  for Computational Linguistics (Volume 2: Short Papers)}, pages 444--449.
  Association for Computational Linguistics.

\end{thebibliography}

\end{document}